\documentclass[sigconf]{acmart}
\AtBeginDocument{%
  \providecommand\BibTeX{{%
    \normalfont B\kern-0.5em{\scshape i\kern-0.25em b}\kern-0.8em\TeX}}}

\setcopyright{acmcopyright}
\copyrightyear{2018}
\acmYear{2018}
\acmDOI{10.1145/1122445.1122456}

\acmConference[Woodstock '18]{Woodstock '18: ACM Symposium on Neural
  Gaze Detection}{June 03--05, 2018}{Woodstock, NY}
\acmBooktitle{Woodstock '18: ACM Symposium on Neural Gaze Detection,
  June 03--05, 2018, Woodstock, NY}
\acmPrice{15.00}
\acmISBN{978-1-4503-XXXX-X/18/06}

\usepackage{mathrsfs}
\usepackage{cleveref}
\usepackage{float}
\usepackage{subfigure}
\usepackage{diagbox}
\usepackage{CJKutf8}
\begin{document}

\title{Sentence-level Online Handwritten Chinese Character Recognition}

\author{Yunxin Li}
\affiliation{%
  \institution{Harbin Institute of Technology}
  \streetaddress{Pingshan 1th Rd}
  \city{Shenzhen}
  \country{China}
  }
\email{liyunxin987@163.com}
\author{Qian Yang}
\affiliation{%
  \institution{Harbin Institute of Technology}
  \streetaddress{Pingshan 1th Rd}
  \city{Shenzhen}
  \country{China}}
\email{20s051056@stu.hit.edu.cn}

\author{Qingcai Chen}
\affiliation{%
  \institution{Harbin Institute of Technology}
  \streetaddress{Pingshan 1th Rd}
  \city{Shenzhen}
  \country{China}}
\email{}
\author{Lin Ma}
\affiliation{%
  \institution{Meituan}
  \streetaddress{Pingshan 1th Rd}
  \city{Beijing}
  \country{China}}
\email{forest.linma@gmail.com}
\author{Baotian Hu}
\affiliation{%
  \institution{Harbin Institute of Technology}
  \streetaddress{Pingshan 1th Rd}
  \city{Shenzhen}
  \country{China}}
\email{hubaotian@hit.edu.cn}
\author{Xiaolong Wang}
\affiliation{%
  \institution{Harbin Institute of Technology}
  \streetaddress{Pingshan 1th Rd}
  \city{Shenzhen}
  \country{China}}
\email{}
\author{Yuxin Ding}
\affiliation{%
  \institution{Harbin Institute of Technology}
  \streetaddress{Pingshan 1th Rd}
  \city{Shenzhen}
  \country{China}}
\email{}



\begin{abstract}
Single online handwritten Chinese character recognition~(single OLHCCR) has achieved prominent performance. However, in real application scenarios, users always write multiple Chinese characters to form one complete sentence and the contextual information within these characters holds the significant potential to improve the accuracy, robustness and efficiency of sentence-level OLHCCR. In this work, we first propose a simple and straightforward end-to-end network, namely vanilla compositional network~(VCN) to tackle the sentence-level OLHCCR. It couples convolutional neural network with sequence modeling architecture to exploit the handwritten character's previous contextual information. Although VCN performs much better than the state-of-the-art single OLHCCR model, it exposes high fragility when confronting with not well written characters such as sloppy writing, missing or broken strokes. To improve the robustness of sentence-level OLHCCR, we further propose a novel deep spatial-temporal fusion network~(DSTFN). It utilizes a pre-trained autoregresssive framework as the backbone component, which projects each Chinese character into word embeddings, and integrates the spatial glyph features of handwritten characters and their contextual information multiple times at multi-layer fusion module. We also construct a large-scale sentence-level handwriting dataset, named as CSOHD to evaluate models. Extensive experiment results demonstrate that DSTFN achieves the state-of-the-art performance, which presents strong robustness compared with VCN and exiting single OLHCCR models. The in-depth empirical analysis and case studies indicate that DSTFN can significantly improve the efficiency of handwriting input, with the handwritten Chinese character with incomplete strokes being recognized precisely.

\end{abstract}





\maketitle
\section{Introduction}

\begin{figure}[t]
    \centering
    \subfigure[Each handwritten character is predicted independently.]{
    \label{sub.1}
    \includegraphics[width=0.43\textwidth]{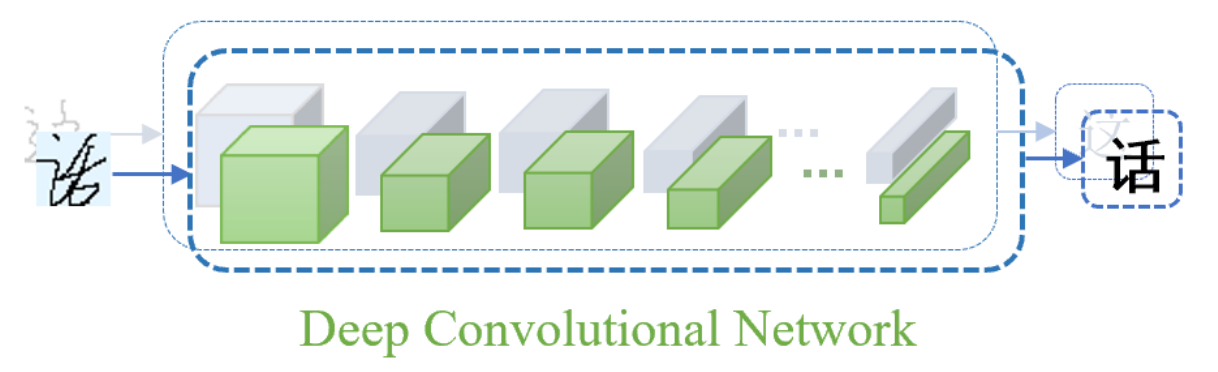}
    }
    
    \subfigure[Handwritten characters have their contextual information.]{
    \label{sub.2}
    \includegraphics[width=0.43\textwidth, height=0.20\textwidth]{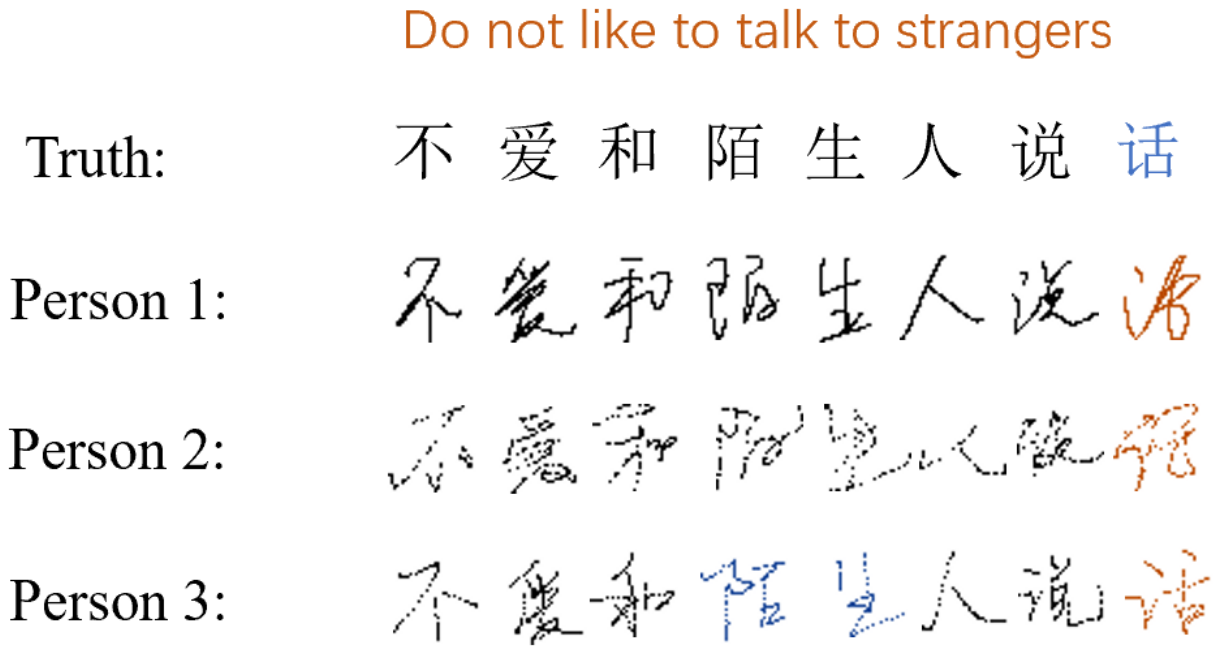}
    }
    \caption{Figure~(a) represents the single online handwritten character recognition method. Figure~(b) represents some handwritten sentences and the blue colored characters are with missing or broken strokes.}
    \label{fig:stroke_com}
\end{figure}

\begin{CJK}{UTF8}{gbsn}
Online handwritten Chinese character recognition~(OLHCCR) has made a  remarkable progress. Most researches~\cite{liu2006online, onlineandoffline, lai2017toward, naik2017online} focus on the single handwritten Chinese character recognition where each character is predicted independently without considering its previous contextual information as shown in Figure~\ref{sub.1}. How to deal with the the sloppy writing, missing or broken strokes is always a challenging problem for OLHCCR. As shown in Figure~\ref{sub.2}, '话'~(speech) written by Person 1 (the last handwritten character) is extremely similar to the character '活'~(live) because of the sloppy writing of '讠' and '氵'. Our preliminary investigation shows that even the state-of-the-art single online handwritten Chinese character recognition model SSDCNN~\cite{Liu_xin} fails to recognize it. As we know, users always write consecutive Chinese characters to form complete sentences in real application scenarios. It is very obvious that '活'~(live) does not conform with the contextual information of the sentence. It indicates that the contextual information has the significant potential to improve the accuracy and robustness of OLHCCR. 
\end{CJK}

For sentence-level online handwritten Chinese character recognition~(sentence-level OLHCCR), the contextual information can also help infer Chinese characters with incomplete strokes to promote the overall efficiency of handwriting input, yet it is a very challenging task. Firstly, it mainly confronts two technical challenges: 1) how to integrate the glyph feature of handwritten character and its previous contextual information. 2) how to design precise, robust, and efficient sentence-level OLHCCR models.
Secondly, at the data level, there is no large-scale Chinese sentence-level handwriting dataset and constructing one needs enormous time and resources.



To tackle the technical challenges of sentence-level OLHCCR, the traditional method considers the contextual information of han-dwritten Chinese characters based on the recognized candidate characters~\cite{wang2012approach_sentences_recognition, quiniou2009word}, which is a multi-stage pipeline instead of the unitized end-to-end approach. The procedure is complicated and inefficient. We argue it can be by natural end-to-end compositional networks and first propose a simple and straightforward end-to-end network, namely vanilla compositional network~(VCN). It couples deep convolutional network with sequence modeling structure such as recurrent neural networks~\cite{lstm, nlm_survey,a_graves_rnn_g} and Transformer\\\cite{attention_is_all}. It integrates the previous contextual information based on the representation of each handwritten character yielded by the deep convolutional network. Although VCN incorporates the contextual information of handwritten characters, it exposes high fragility while confronting with not well written characters, as the convolutional representations of them are indistinguishable due to the irregular shapes of them.

To further enhance the robustness of end-to-end network, we design the deep spatial-temporal fusion network~(DSTFN) that contains the core multi-layer fusion module and glyph representation module. We adopt word embeddings to represent the previously recognized handwritten Chinese characters while predicting at ea-ch time step, which overcomes the problem of inadequate representations of not well written characters. Specifically, we utilize a Transformer based pre-trained autoregressive framework as the backbone structure. The representation of handwritten Chinese characters obtained by the glyph representation module and their contextual information are integrated multiple times in the multi-layer fusion module. Such proposed multi-layer fusion method that continuously integrates contextual information and glyph features makes the architecture of DSTFN stable.

Moreover, we reorganize a large-scale Chinese sentence-level online handwriting dataset named as CSOHD\footnote{We will open it in the future.}. We construct it by the way of pair based on collected approximately 2.4 million sentences from Chinese Wikipedia\footnote{https://dumps.wikimedia.org/zhwiki/} and Sougou news corpus\footnote{https://www.sogou.com/labs/resource/list\_yuliao.php} and 900 sets of single handwritten Chinese character. We conduct extensive experiments on CSOHD and the results demonstrate the performance of DSTFN significantly outperforms the exiting single OLHCCR model and VCN. And the in-depth empirical analysis indicates that DSTFN is more robust and efficient compared with other models, recognizing the handwritten Chinese character with incomplete strokes precisely. 



The main contributions of our work are three-fold:
\begin{itemize}
    \item We present two end-to-end models for sentence-level OLHCCR: vanilla compositional network~(VCN) and deep spa-tial-temporal fusion network~(DSTFN) that integrates the glyph feature of handwritten character and its previous contextual information.   
    \item To address the lack of large-scale handwritten sentences, we collect enormous sentences and handwritten character sets to construct a Chinese sentence-level online handwriting dataset, namely CSOHD.
    \item Extensive experiments demonstrate that DSTFN is more precise, robust and efficient than VCN and the state-of-the-art single OLHCCR model on the constructed CSOHD dataset. 
\end{itemize}


\section{Related Works} 

In this section, we introduce the recent methods of OLHCCR 
and multi-modal fusion.


In the early researches of handwritten Chinese character recognition, researchers focus solely on the structural feature of Chinese characters. With the advent of statistical learning methods, most of handwritten Chinese character recognition are mainly divided into structural methods and statistical methods~\cite{two_methods,DirMap}, and the above two methods mainly rely on the handcraft features. Recently, the deep conventional neural networks~(DCNNs) such as ZFNet~\cite{zfnet}, GoogLeNet~\cite{googlenet}, VGG~\cite{vgg} and ResNet~\cite{icpr}, can capture the complex features of images and perform well on many image tasks such as digital classification~\cite{byerly2020branching}, image classification~\cite{cnn_image} and image representation~\cite{convolutional_for_image_representation}. Hence, researchers utilize DCNNs~(e.g. MCDNN~\cite{MCDNN}, SSDCNN~\cite{Liu_xin}) to learn multi-dimension features of handwritten Chinese characters, which achieves significant success~\cite{2013ICDAR,directional_feature_2,directional_feature_3, locality}. The above-mentioned traditional methods usually address issues in three stages: data processing, feature extraction, and classification~\cite{Liu_xin}.

For OLHCCR, \citet{Liu_xin} propose SSDCNN model that integrates the sequence of strokes and eight directional features of Chinese characters, achieving the state-of-the-art performance on single Chinese character recognition. However, the application scenario of OLHCCR is generally at sentence level. The single online handwritten Chinese character regonition~(single OLHCCR) model and traditional sentence-level handwriting recognition models~\cite{wang2012approach_sentences_recognition} confront the intrinsic difficulty of inferior efficiency compared to the sentence-level pinyin input method~\cite{pinyin}. To step forward the research of sentence-level, We first reorganize a novel dataset by the way of pair based on millions of sentences from large news corpus and hundreds of handwritten sets of single Chinese character, to simulate the sentence-level handwriting input scene. 


Compared to single OLHCCR, handwritten Chinese characters of sentence-level OLHCCR contain the contextual information as well as the static glyph features of them. It can be reduced to the sequential classification task according to the step-by-step inferring way. As we know, for the similar natural language text generation task such as Abstractive Summarization~\cite{bottom_abstractive_summary, cnn-daily}, Question Generation ~\cite{squad-2018} and Paraphrase Generation~\cite{paraphrse_pair_wise}, recurrent neural networks~(RNNs)~\cite{survey_of_nlg, lstm} and Transformer~\cite{attention_is_all} based models~\cite{GPT-2, xlnet, T5, bart_original} can encode the natural language text and generate coherent text. Hence, we apply language models to capture the contextual information of handwritten characters and first propose a simple vanilla compositional network~(VCN).

While confronting with not well written characters, the representation of handwritten Chinese characters obtained by deep convolutional networks is insufficient. However, the word representation methods~(e.g. one-hot encoding, distributed representation~\cite{distributed_word}) can convert any character into uniquely identified word embeddings without considering the glyph feature of it~\cite{glyce_ljw} and have make a remarkable progress in natural language processing. Hence, we utilize word embeddings to represent previously recognized handwritten Chinese characters and propose a deep spatial-temporal fusion network~(DSTFN). It integrates the contextual information encoded in word embeddings space and the extracted glyph representation of sequential handwritten characters, containing two-modal features. Inspired by the previous fusion methods of multi-modal representation such as Early Fusion LSTM~\cite{williams-etal-2018-recognizing}, Tensor Fusion Network~\cite{zadeh-etal-2017-tensor}, Memory Fusion Network~\cite{zadeh2018memory} and Low-rank multi-modal Fusion~\cite{lowrank}, we design a multi-modal fusion sub-layer in each block of DSTFN to tackle the above features.  
\section{Our Methodology}
We introduce the problem formulation in Section~\ref{tf} and the vanilla compositional framework in Section~\ref{sec_sp}. Then, we propose the deep spatial-temporal fusion network in Section~\ref{sec_dstfrn}.

\subsection{Problem Formulation}
\label{tf}

 Compared to single OLHCCR whose input and output both are single character, the input and output of sentence-level OLHCCR are sentences. The input is the sentence composed of consecutive handwritten Chinese characters and each character consists of multiple strokes, denoted as $\mathbf{H}=(s_{1,1}, ..., s_{i,j},..., s_{i,k_{i}},..., s_{n,k_{n}})$, where $s_{i,j}$ represents the $j$th stroke of the $i$th handwritten character, $s_{i,k_{i}}$ represents the final stroke of $i$th handwritten character and $n$ is the number of characters. Despite users do not necessarily write Chinese characters in the correct order of strokes, each handwritten Chinese character can be divided into multiple fragments~(at least the whole character) according to the handwriting pattern of humans. The output is the standard Chinese sentence denoted as $\mathbf{T}=({w}_{1},...,{w}_{n})$, where ${w}_{i}$ represents the $i$th word in the sentence.
 

\begin{figure}[t]
    \centering
    \includegraphics[width=0.45\textwidth]{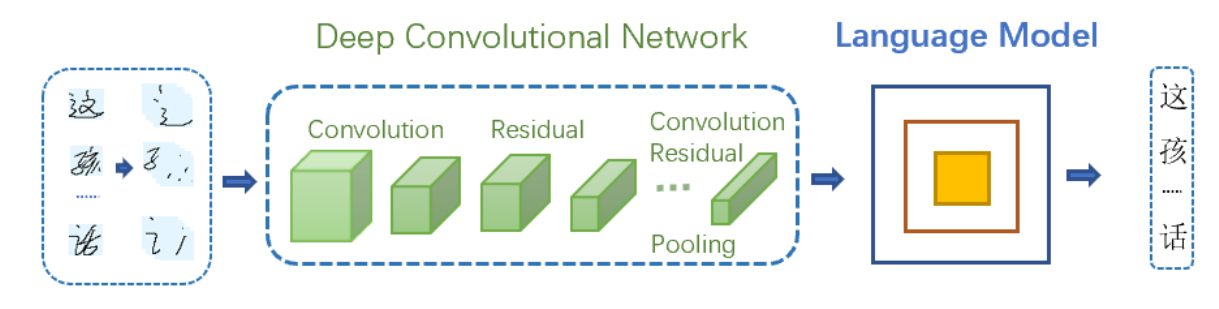}
    \caption{The overall structure of vanilla compositional network composed of deep convolutional network and language model.}
    \label{fig:model_1}
\end{figure}

\subsection{Vanilla Compositional Network}
\label{sec_sp}

For sentence-level OLHCCR, an intuitive approach is to use deep convolutional neural network for single handwritten Chinese character representation with one followed language model to capture the previous contextual information of it as shown in Figure~\ref{fig:model_1}. We name this simple end-to-end framework vanilla combination network~(VCN). Concretely, we formulate the strokes based deep convolutional neural network~(SBDCNN) to extract the image features of handwritten Chinese characters, which utilizes the method presented in state-of-the-art single OLHCCR model SSDCNN~\cite{Liu_xin}. 
Then, we apply language models LSTM~\cite{lstm} or GPT-2~\cite{GPT-2} to learn the contextual information of handwritten characters, i.e. the dependency patterns between characters in handwritten sentences. Note that each character only observe its previous characters. The whole process is that the sequential representation obtained by SBDCNN is inputted into the downstream language model to get the hidden state that incorporates the contextual information. Finally, we utilize a Linear and Softmax layer as the final output layer of VCN to predict the target words of input sequential handwritten Chinese characters. 


\subsection{Deep Spatial-Temporal Fusion Network}
\label{sec_dstfrn}

Although VCN incorporates the contextual information of handwritten characters, the convolutional representation of characters with not well written shape~(e.g. sloppy writing, missing or broken strokes) is inadequate to be directly used in the downstream network because the representation of them is indistinguishable and it easily causes cascading errors while inferring.



To address the inadequate representation problem of not well written characters and deeply integrate glyph features of handwritten characters and their contextual information, we propose the deep spatial-temporal fusion network~(DSTFN). As shown in Figure~\ref{fig:model_2}, DSTFN is composed of core \emph{multi-layer fusion module} and \emph{glyph representation module}. We introduce the structure and function of each module in detail below.



\subsubsection{\textbf{Glyph Representation Module}} 
\begin{figure}[t]
    \centering
   \includegraphics[width=0.450\textwidth, height=0.50\textwidth]{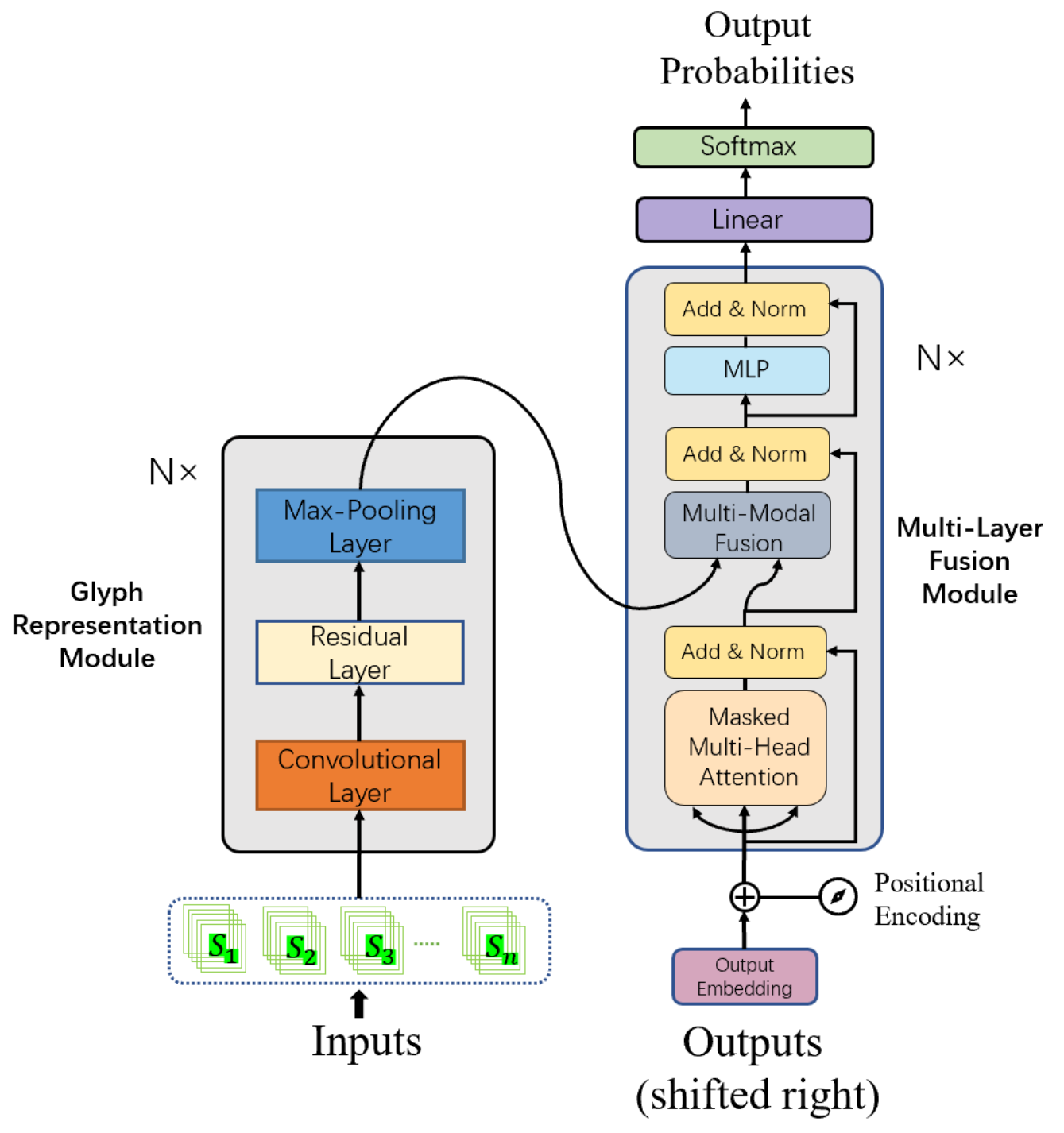}
    \caption{Overview of the Deep Spatial-Temporal Fusion Network for sentence-level OLHCCR. ($\mathbf{S}_1, \mathbf{S}_2, ..., \mathbf{S}_n$) represent the sequential multi-channel stroke maps of handwritten characters. The glyph representation module is used to represent the handwriting input. The multi-modal fusion sub-layer is used to integrate the glyph feature of handwritten characters and their contextual information gained by the masked multi-head attention sub-layer. The value of N is set to 4.}
   \label{fig:model_2}
\end{figure}

The glyph representation module is used to obtain the sequential representation of handwritten characters. We utilize the aforementioned SBDCNN model as the glyph representation framework except for the different output dimension of the last fully connected layer as shown in Figure~\ref{fig:model_2}. We set the image size of each stroke to 32$\times$32 and the maximum number of each character to 28. More concretely, for any handwritten Chinese character, we fill the gap between the last stroke and the maximum number set with full-zero feature maps. The position on each feature map is set to 1 where the strokes pass; otherwise, the position is set to 0.  

The architecture of glyph representation module contains four identical blocks and each block has three sub-layers, which are the convolutional layer, residual layer and max-pooling layer in turn. The multi-channel stroke maps encoded by convolutional sub-layer of the $l$th layer can be denoted as Eq.~\ref{eq3}.
\begin{equation}
\mathbf{z}_{c}^{l} = \mathbf{ReLU}(\mathbf{W^{l}}\mathbf{z}^{l-1} + \mathbf{b^{l}})
\label{eq3}
\end{equation}
where $\mathbf{z}^{l-1}$ is the output of the $l-1$th layer. $\mathbf{w}^{l}$ and $\mathbf{b}^{l}$ are the parameters of the convolutional layer. The output $\mathbf{z}_{c}^{l}$ then passes the residual network layer~\cite{icpr} and the process can be denoted as Eq.~\ref{eq4}.
\begin{equation}
    \mathbf{z}_{r}^{l} = \mathbf{W}_{s}^{l}\mathbf{z}_{c}^{l} + \mathcal{F}(\mathbf{z}_{c}^{l}, \mathbf{W}_{r}^{l})
    \label{eq4}
\end{equation}
    where the function $\mathcal{F}(\mathbf{z_{c}}^{l}, \mathbf{W}_{r}^{l})$ represents the residual operation block. It generally contains the convoluitonal layer and activation function. Note that we perform the projection parameter $\mathbf{W}_{s}^{l}$ for the shortcut connection to make sure that the output and input of residual block have the matching dimensions.
The residual operation layer is followed by the max-pooling operation, which can expand the horizon of perception and maintain the invariance of spatial rotation and translation. We utilize 2$\times$2 window for multi-channel feature maps $\mathbf{z}_{r}^{l}$, which can be denoted as Eq.~\ref{eq5}. 
\begin{equation}
   \mathbf{z}^{l} = \mathbf{Max}_{ 2\times2}(\mathbf{z}_{r}^{l}) 
   \label{eq5}
\end{equation}
The whole process is iterated through four convolutional blocks until gaining the fixed length output vector. We can gain the final representation of one handwritten Chinese character, denoted as $\mathbf{R}=(r_1,..., r_m)$ , where $m$ is the dimension of the output vector. Hence, the representation of sequential handwritten Chinese characters can be denoted as $\mathcal{R}=(\mathbf{R}_1, \mathbf{R}_2,...,\mathbf{R}_n)$ where $n$ is the length of the input sequence.

\subsubsection{\textbf{Multi-layer Fusion Module}} 
After obtaining the glyph re-presentation of handwritten characters, we utilize the multi-layer fusion module shown in Figure~\ref{fig:model_2} to integrate the glyph feature of handwritten characters and their contextual information.

First, we project the sentence example $\mathbf{T}=({w}_{1},...,{w}_{n})$ into the fixed-dimensional word embeddings via a learnable embedding table, which has the start symbol $BOS$. Meanwhile, we incorporate the learned positional information for the sequence as shown in Figure~\ref{fig:model_2} to help learn the dependency relationship between words. After the word representation is obtained, we first pre-train Transformer decoder based auoregressive framework by the task of predicting the next word, where each layer contains two sub-layers, i.e. Masked Multi-Head Attention layer and Multi Perceptron layer. It can obtain context-sensitive word representation and generate coherent sentences. Ideally, the pre-trained auoregressive framework can capture the dependency between words and predict the following word accurately, given the preceding words. 

However, it is notable that sentence-level OLHCCR has continuous strokes input of the target word at each time step of predicting besides the outputs of previous steps compared to text generation task~\cite{2020_text_generation}. Inspired by the above observation, we incorporate the prompting glyph information of the target word into each layer of the aforementioned autoregressive framework by adding the extra multi-modal fusion sub-layer, and the other sub-layers are identical as the structure of pre-trained autoregressive framework. The overall multi-layer fusion module has four identical blocks as shown in Figure~\ref{fig:model_2}. Each block can be computed as Eq.~\ref{eq9}.


\begin{equation}
\begin{array}{c}
\mathbf{h}_{l}^{M}=\mathbf{LayerNorm}(\mathbf{h}_{l-1} + \mathbf{MultiHeadAttention}(\mathbf{h}_{l-1}))\vspace{1ex} \\
\mathbf{h}_{l}^{F} = \mathbf{LayerNorm}(\mathbf{h}_{l}^{M} + \mathbf{MultiModalFusion}(\mathbf{h}_{l}^{M})) \vspace{1ex}\\
\mathbf{h}_{l} = \mathbf{LayerNorm}(\mathbf{h}_{l}^{F} + \mathbf{MLP}(\mathbf{h}_{l}^{F}))  
\end{array}
\label{eq9}
\end{equation}
where $\mathbf{h}_{l-1}$ is the output of the $l-1$th layer. $\mathbf{MultiHeadAttention}$ is based on the multi-head attention mechanism, which can capture the contextual information of each handwritten character. $\mathbf{MLP}$ is composed of fully connected neural network and activation function as the feed-forward networks of Transformer~\cite{attention_is_all}. The computation of  $\mathbf{MultiModalFusion}$ is denoted as Eq.~\ref{eq10}.
\begin{equation}
    \mathbf{y} = \mathbf{W}_{F}^{l}[\mathbf{h}_{l}^{M},\mathcal{R}] + \mathbf{b}^{l}
    \label{eq10}
\end{equation}
where $[,]$ represents the concatenate operation and $\mathbf{W}_{F}^{l}$, $\mathbf{b}^{l}$ are the projection parameters. Concatenate operation is a useful method for integrating two modal information generally~\cite{williams-etal-2018-recognizing}. The top output of multi-layer fusion module, denoted as $\mathbf{G}=(\mathbf{g}_1,...,\mathbf{g}_n)$, goes through a Linear and Softmax layer to obtain the final prediction probability as Eq.~\ref{eq7}.
\begin{equation}
    P(\mathbf{x}=(\mathbf{x}_1,...,\mathbf{x}_n)) = \mathbf{Softmax}(\mathbf{W}\mathbf{G} + \mathbf{b})
    \label{eq7}
\end{equation}
where $P(\mathbf{x}=(\mathbf{x}_1,...,\mathbf{x}_n))$ is the prediction probability of the whole sequence of input handwritten Chinese characters and $\mathbf{W}$, $\mathbf{b}$ are the parameters.

To summarize, the method of utilizing word embeddings to represent identified handwritten characters enriches the representation of them and further addresses the inadequate representation problem of not well written characters. The multi-layer fusion module integrates the contextual information of handwritten Chinese characters and the glyph features of them multiple times, to highly integrate the temporal and spatial features. The whole architecture of DSTFN is more stabilized and stronger compared to VCN.


\begin{figure}[t]
    \centering
    \includegraphics[width=0.45\textwidth]{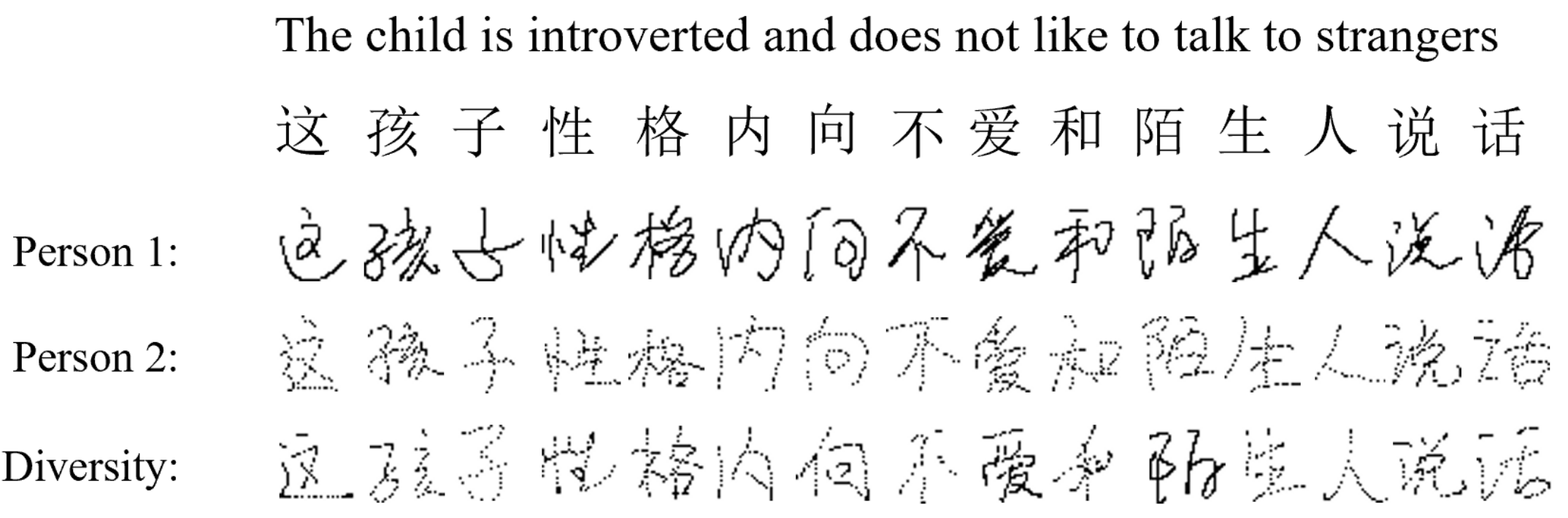}
    \caption{Person 1/2 represent each character is from the same handwritten set. Diversity represents each character is from different handwritten sets.}
    \label{fig:example_dataset}
\end{figure}

\subsection{Training and Inference} For training, suppose the target sequence $\mathbf{T}=({w}_{1},...,{w}_{n})$ has been obtained, VCN can be trained via minimizing the loss function as Eq.~\ref{eq8}.
\begin{equation}
    \mathcal{L}_{sg} = -\sum_{i=1}^{n}\mathbf{log}{P}_{i}(x_i={w}_{i}|s_{1,1}, ..., s_{i,j}, ..., s_{i,k_{i}})
    \label{eq8}
\end{equation} and DSTFN can be trained via minimizing the loss function as Eq.~\ref{eq11}.
\begin{equation}
    \mathcal{L}_{DFTRN}= -\sum_{i=1}^{n}\mathbf{log}{P}_{i}(x_i={w}_{i}|s_{1,1}, ..., s_{i,j}, ..., s_{i,k_{i}}; w_1,..., w_{i-1})
    \label{eq11}
\end{equation}
where $(s_{1,1}, ...., s_{i,j}, ..., s_{i,k_{i}})$ represent the sequential handwritten Chinese characters and $(w_1,...,w_{i-1})$ represent the target sequence of input handwritten characters before the $i$th position.

For inferring, at each time step, the input of VCN is the sequential handwritten characters until the current time step, yet the input of DSTFN has the characters recognized in the previous time steps besides the above inputs. The outputs of the two methods are both the recognition results of all handwritten Chinese characters up to the current time step.

\section{Chinese sentence-level online handwriting Dataset}


\label{data_p}

\subsection{Dataset Construction}
\textbf{Handwritten Chinese Character Set.} The character set is 3755 Chinese characters based on the level-1 of GB2312-80 and we collect 900 sets of handwritten Chinese characters from the exiting data set such as CASIA-OLHWDB 1.0 and 1.1, HIT-OR3C and SCUT-COUCH2009, where only 513 sets of them contain the complete 3755 Chinese characters. Hence, We select 200 complete sets of handwritten Chinese characters for testing and select another 100 ones for validation set. The rest 600 sets of handwritten characters are utilized for training.  

\noindent\textbf{Sentence Corpus.} We collect enormous texts from Sougou news corpus and Chinese Wikipedia, and then split them into sentences according to segmentation symbols. We clear up the special symbols in the sentences and filter out the sentences that have characters not contained in the aforementioned character set. We retain sentences between 10 and 40 words. The detailed statistics are shown in Table~\ref{tab:data_stat}.

\noindent\textbf{Reorganization.} We match the Chinese characters in every sentence with the corresponding handwritten forms to reorganize a large-scale Chinese sentence-level online handwriting dataset na-med as CSOHD. As 'Diversity' shown in Figure~\ref{fig:example_dataset}, each Chinese character has various matching handwritten forms, so each sentence can have millions of handwriting styles because each Chinese character has approximately 900 handwritten samples. Such simple yet effective way of reorganization can simulate almost any online handwriting scenario, based on large-scale sentence corpus and various handwritten Chinese characters.       

\subsection{Data Statistics}

\begin{table}[t]
    \centering
     \caption{The detailed statistics of CSOHD. Len-1/2/3(\%) represent the proportion of the three levels of sentence length, which is divided according to an interval of 10 words.}
    \label{tab:data_stat}
    \begin{tabular}{lcccc}
    \hline
    \multicolumn{1}{c}{\textbf{CSOHD}} & \multicolumn{1}{c}{\textbf{Size}} & \multicolumn{1}{c}{\textbf{Len-1(\%)}} & \multicolumn{1}{c}{\textbf{Len-2(\%)}} &
    \multicolumn{1}{c}{\textbf{Len-3(\%)}}\\
    \hline
    \# Train & $2,386,485$ & $31.63$ & $34.58$ & $32.79$\\
    \# Dev & $15,412$ & $23.60$ & $42.89$ & $33.51$\\
    \# Test & $46,089$ & $20.31$ & $41.80$ & $37.89$\\ \hline
    \end{tabular}
\end{table}

\begin{figure}[t]
    \centering
    \subfigure[Train set]{
    \label{w.1}
    \includegraphics[width=0.22\textwidth]{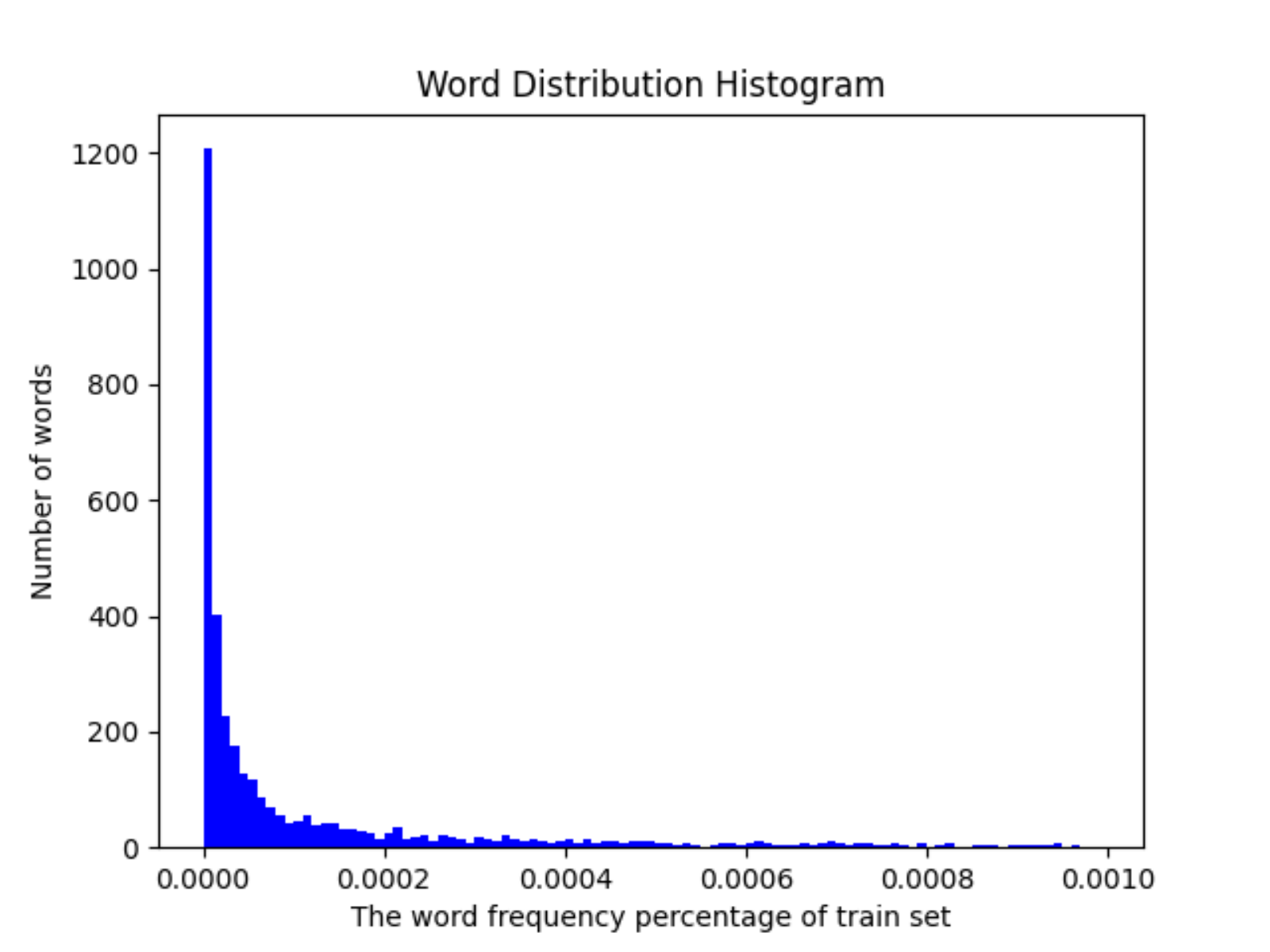}
    }\subfigure[Dev set]{
    \label{w.2}
    \includegraphics[width=0.22\textwidth]{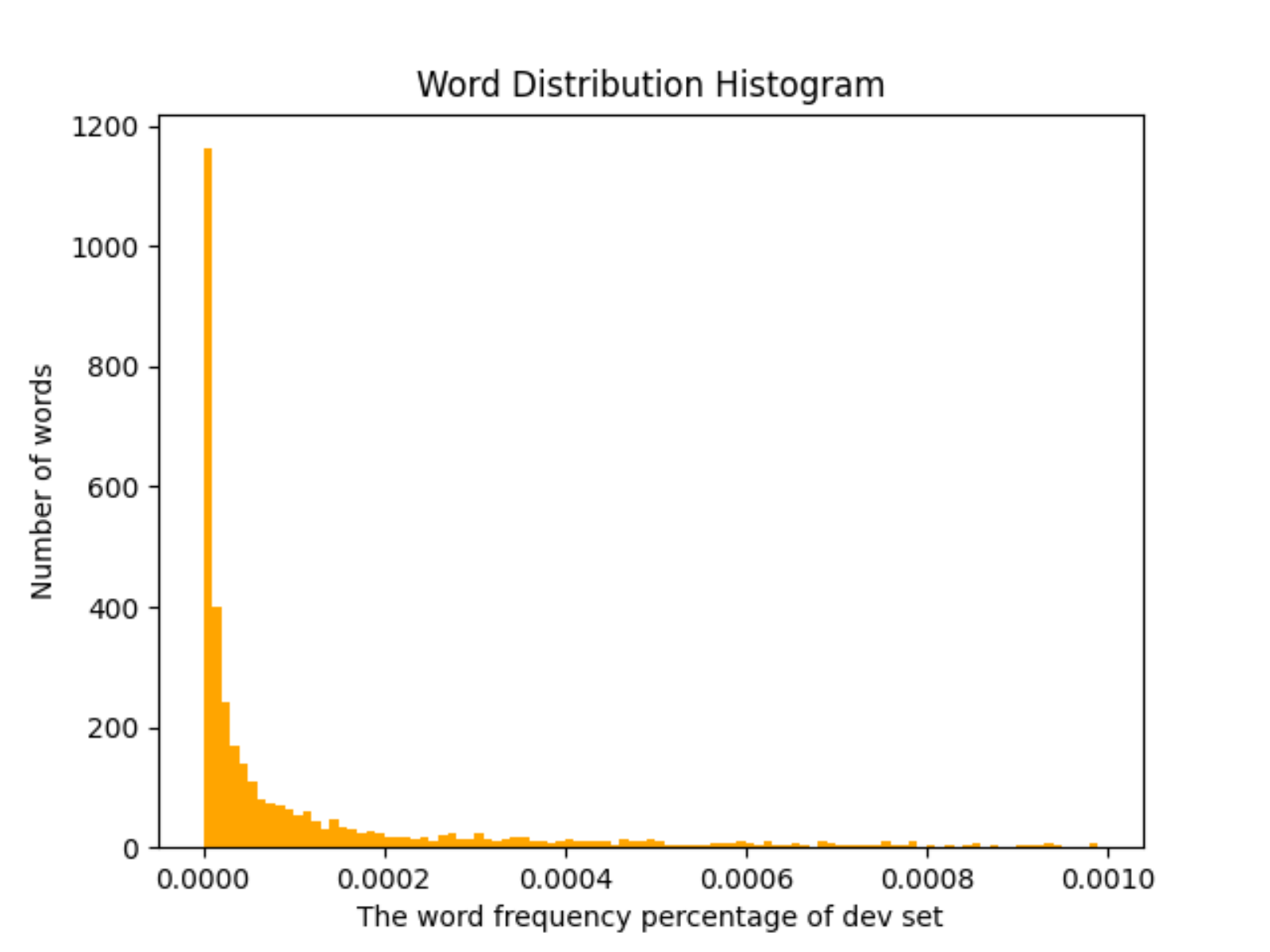}
    }
    
    \subfigure[Test set]{
    \label{w.3}
    \includegraphics[width=0.22\textwidth]{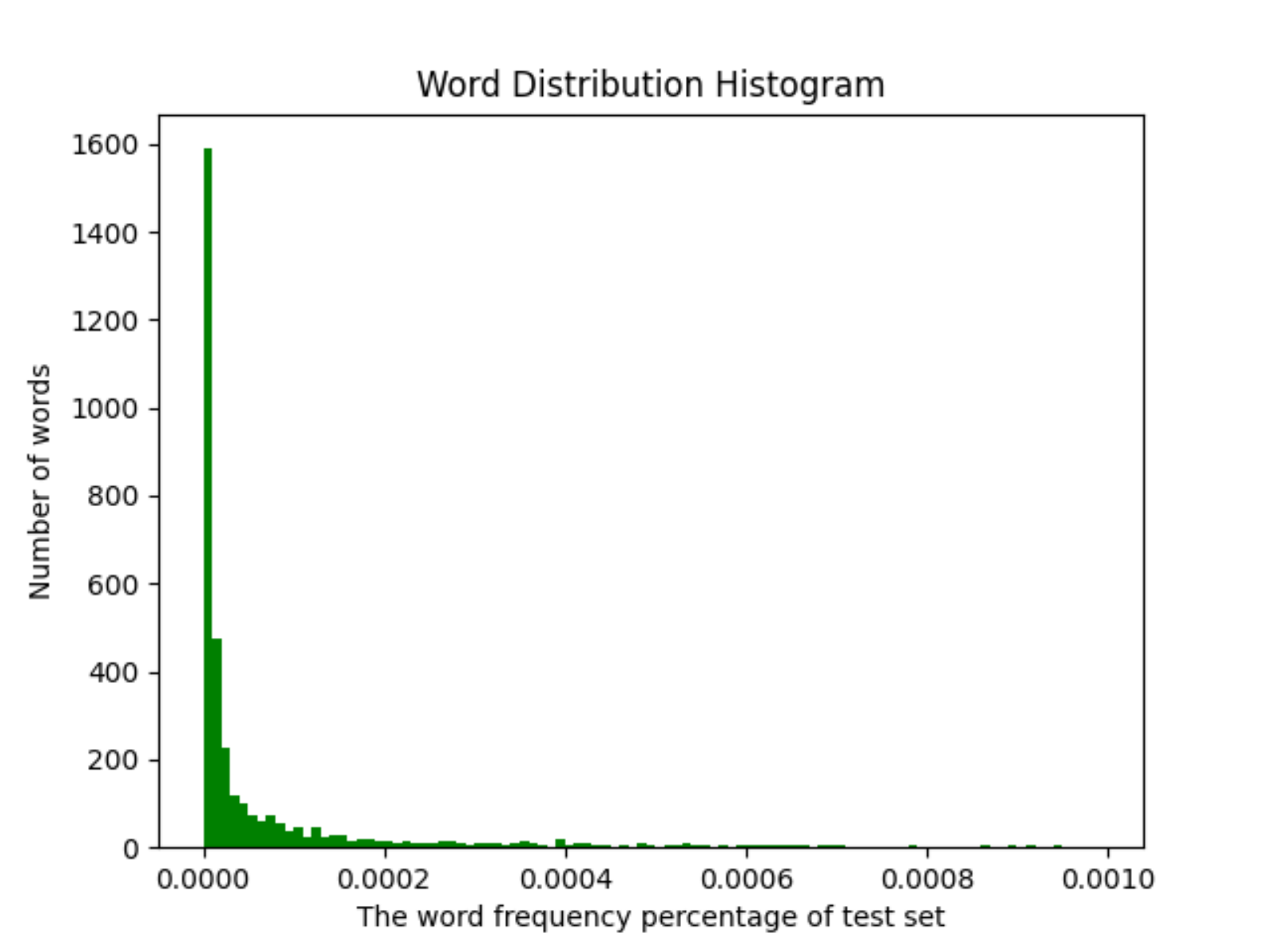}
    }\subfigure[Statistics of sentence length ]{
    \label{w.4}
    \includegraphics[width=0.21\textwidth, height=0.155\textwidth]{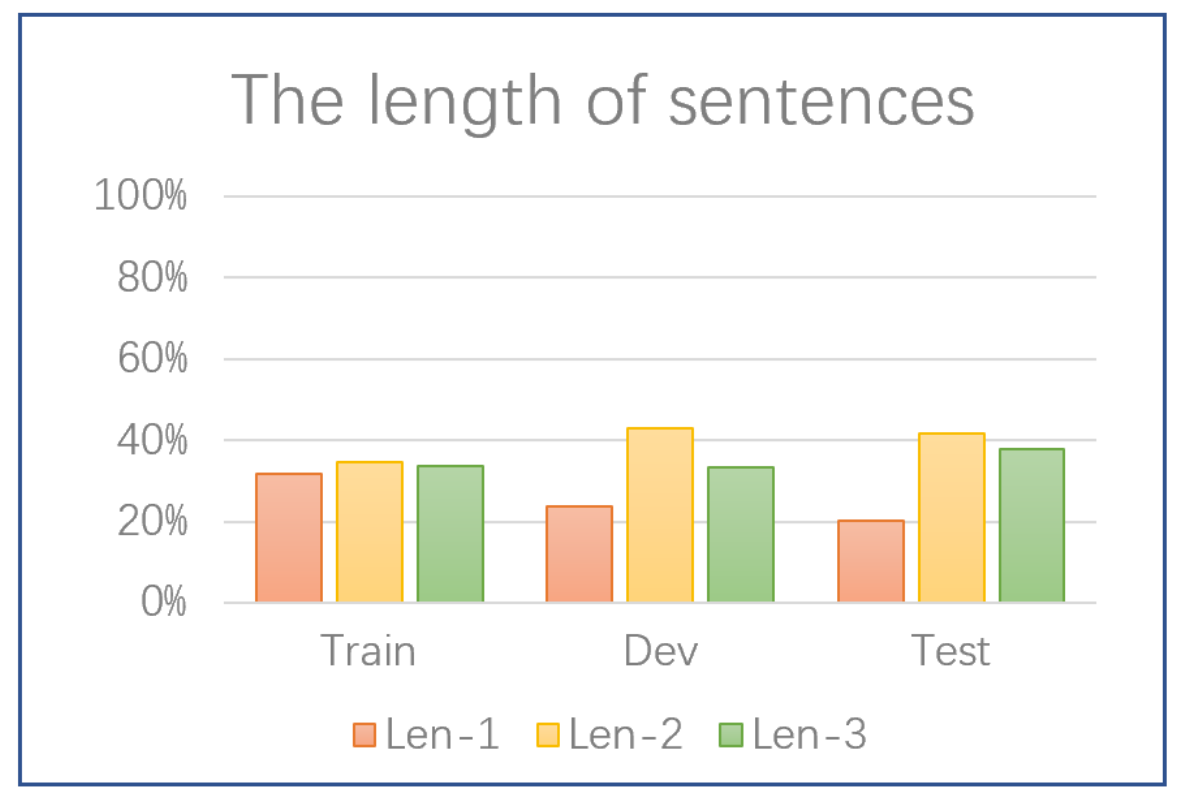}
    }
    \caption{The statistics of SOHCD, containing the word distribution of dataset and the distribution of sentence length.}
    \label{fig:w_s}
\end{figure}

\noindent\textbf{Data Analysis.} As shown in Table~\ref{tab:data_stat} and Figure~\ref{w.4}, we count the proportion of sentence length in the dataset, where the length of the sentence is divided into three levels according to the interval of 10 words. To better evaluate the recognition ability of models at sentence level, sentences with more than 20 Chinese words in the validation and test set account for more proportion, compared to the uniform distribution of the three length levels in the training set. Note that the $3755$ Chinese characters must be contained in the train/dev/test set in order to fully evaluate all handwritten Chinese characters. Moreover, from Figure~\ref{w.1} to Figure~\ref{w.3}, we observe that the word distribution of train, dev and test set are almost identical, which ensure the consistency of data features to a certain extent. Most words are low-frequency and a few commonly used words are high-frequency, which conforms to the human habit of using Chinese. 
\begin{table}[t]
    \centering
    \caption{Automatic evaluation on the test set under the condition that all handwritten Chinese characters has the full strokes. P@i (i=1,2,3,4,5) represents the proportion of the i results with the highest probability that contains the correct category.}
    \label{tab:result_1}
    \begin{tabular}{lccccc}
         \multicolumn{1}{c}{\textbf{Model}} & \multicolumn{1}{c}{\textbf{P@1}} & \multicolumn{1}{c}{\textbf{P@2}} & \multicolumn{1}{c}{\textbf{P@3}} & \multicolumn{1}{c}{\textbf{P@4}}
         & \multicolumn{1}{c}{\textbf{P@5}}\\\bottomrule
         SBDCNN & $97.68$& $98.69$& $98.97$ & $99.16$& $99.23$\\ \hline
         VCN (LSTM) & $97.99$& $98.72$& $98.96$ & $99.10$ &$99.17$\\ \hline
         VCN (GPT)& $98.62$& $99.31$& $\mathbf{99.49}$ & $\mathbf{99.55}$ &$\mathbf{99.61}$\\ \hline
         DSTFN & $\mathbf{98.73}$& $\mathbf{99.33}$& $99.45$& $99.53$ & $99.59$\\ \bottomrule 
    \end{tabular}
\end{table}

\section{Experiments}

\subsection{Experiment Setting}
\label{experi}
\emph{SBDCNN:} there are various single OLHCCR models which have achieved significant performance such as DSamCNN~\cite{dasmple}, DirMap-CNN~\cite{DirMap} and MCDNN~\cite{MCDNN}. We adopt the state-of-the-art model SBDCNN~\cite{Liu_xin} that achieves the high performance and can model the sequence of strokes. \emph{VCN:} we select sequence modeling architectures LSTM and GPT-2 for the downstream network, named VCN~(LSTM) and VCN~(GPT). We set the depth of LSTM and GPT-2 both to 4. \emph{DSTFN:} the dimension of the hidden size is 256, identical as the aforementioned VCN model. Each block of it has eight attention heads. 


We construct our vocabulary on Chinese Internal Code Specification\footnote{http://www.sunchateau.com/free/fantizi/ziku/gbk.htm} whose size is 3755. To speed up training, we load the stroke maps to memory at first instead of repeatedly loading. Due to the limit of the memory size, we load 150 sets each time, and randomly reload the handwritten Chinese character sets every 1000 steps to get new sets. For the selected 150 sets, we randomly select one handwritten sample for each character of the sentence. For VCN and DSTFN, we pre-train LSTM, GPT-2 and Transformer based autoregressive framework on the training sentences for 8 epochs to make them better understand the sentences and the maximum sequence length is set to 512. During training, each character in sentences is represented by one 28 x 32 x 32 tensor and we train the above models on 3 V100 GPUs through the Adam~\cite{adam} optimizer with the base learning rate of 1e-3 and set the batch size to 64.

\subsection{Results and Analysis}
In this section, we present and analyse the performances of models introduced in Section \ref{sec_sp} and \ref{sec_dstfrn} on our designed experiments.

\subsubsection{\textbf{Accuracy}}
\begin{table}[t]
    \centering
    \caption{Automatic evaluation on the test set with five selected sets of handwritten characters. We uniformly utilize p@1 as the evaluation metric. HS-1, 2, ..., 5 represent the selected five sets of single handwritten Chinese character.}
    \label{tab:result_2}
    \begin{tabular}{lccccc}
         \multicolumn{1}{c}{\textbf{Model}} & \multicolumn{1}{c}{\textbf{HS-1}} & \multicolumn{1}{c}{\textbf{HS-2}} & \multicolumn{1}{c}{\textbf{HS-3}} & \multicolumn{1}{c}{\textbf{HS-4}}
         & \multicolumn{1}{c}{\textbf{HS-5}}\\\bottomrule
         SBDCNN & $95.26$& $70.20$& $81.14$ & $90.03$& $97.05$\\ \hline
         VCN (LSTM) & $97.04$& $71.56$& $88.45$ & $91.43$ &$97.03$\\ \hline
         VCN (GPT) & $97.87$& $76.46$& $89.29$ & $91.80$ &$98.25$\\ \hline
         DSTFN & $\mathbf{98.15}$& $\mathbf{78.47}$& $\mathbf{92.25}$& $\mathbf{91.89}$ & $\mathbf{98.32}$\\ \bottomrule 
    \end{tabular}
\end{table}
\begin{figure}[t]
    \centering
    \includegraphics[width=0.40\textwidth]{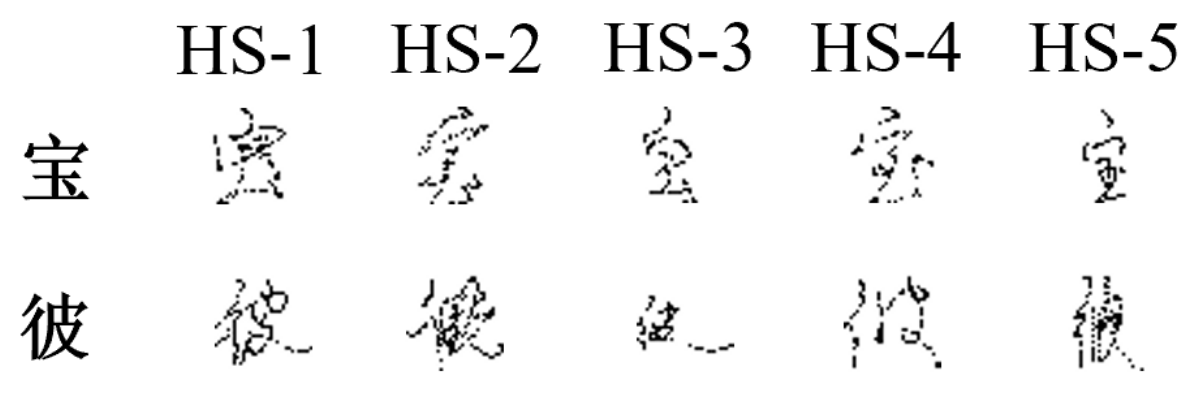}
    \caption{Two handwritten Chinese character samples of the selected five sets. The written shapes of HS-2/3 are more sloppy than others.}
    \label{fig:stroke_two}
\end{figure}


We first test all models under the circumstance that each handwritten character has complete strokes, which can evaluate the best performance of models without considering the robustness and efficiency at sentence level. During inferring, the number of handwritten character sets are too large to be tested for each set. Hence, we randomly extract one handwritten sample from the 200 handwritten sets for each character and test \textbf{five} times for each model to ensure the confidence of experiment results. The average results are shown in Table~\ref{tab:result_1}. Moreover, we select five sets from the total 200 test sets of handwritten Chinese characters and separately test characters with the selected five forms for all models. The experiment result is shown in Table~\ref{tab:result_2}. We also select two handwritten character samples from the selected five handwritten sets, shown in Figure~\ref{fig:stroke_two}, where the handwritten characters of HS-2/3/4 are extremely irregular. Note that the selected five handwritten sets are used in the following experiments.  

The experiment results in Table~\ref{tab:result_1} show that VCN~(GPT) and DSTFN preform well on all evaluation metrics, especially on P@1 where they both exceed SBDCNN by about 1 point. Compared with single-character recognition model, models that consider the previous contextual information of characters perform better on most evaluation metrics. Furthermore, from the experiment results of Table~\ref{tab:result_2}, we observe that the recognition accuracy of DSTFN is very higher than single-character model SBDCNN especially on irregular and scribbled handwritten Chinese characters~(P2: $78.47$ vs $70.20$, P3: $92.25$ vs $81.14$). The above experiment results demonstrate that single-character recognition model can not recognize irregular and scribbled handwritten Chinese characters well without considering their contextual information, and models that incorporate the contextual information of handwritten characters are more robust and generalized. Compared to VCN(LSTM), VCN(GPT) performs better on all evaluation metrics, which indicates that pre-trained language models with huger parameters can learn more abundant sentence-level contextual information. An interesting phenomenon shown in Table~\ref{tab:result_1} is that the recognition accuracy gap between DSTFN and VCN~(GPT) is not great, which may be attributed to the fact that the glyph representation of handwritten Chinese character and its corresponding word embeddings have almost identical effect as the representation of them with complete strokes.  
\begin{table}[]
    \centering
    \caption{Automatic evaluation on the test set under the condition that all handwritten Chinese characters retain 70\% of strokes.}
    \label{tab:result_3}
    \begin{tabular}{lccccc}
         \multicolumn{1}{c}{\textbf{Model}} & \multicolumn{1}{c}{\textbf{P@1}} & \multicolumn{1}{c}{\textbf{P@2}} & \multicolumn{1}{c}{\textbf{P@3}} & \multicolumn{1}{c}{\textbf{P@4}}
         & \multicolumn{1}{c}{\textbf{P@5}}\\\bottomrule
         SBDCNN & $72.21$& $78.94$& $82.11$ & $84.20$& $85.65$\\ \hline
         VCN~(GPT) & $76.10$& $81.21$& $83.57$ & $85.09$ &$86.07$\\ \hline
         DSTFN & $\mathbf{81.97}$& $\mathbf{87.28}$& $\mathbf{89.48}$& $\mathbf{90.71}$ & $\mathbf{91.51}$\\ \bottomrule 
    \end{tabular}
\end{table}
\begin{table}[]
    \centering
    \caption{Automatic evaluation~(P@1) on the test set with five selected handwritten sets where each handwritten Chinese character retains 70\% of strokes.}
    \label{tab:result_4}
    \begin{tabular}{lccccc}
         \multicolumn{1}{c}{\textbf{Model}} & \multicolumn{1}{c}{\textbf{HS-1}} & \multicolumn{1}{c}{\textbf{HS-2}} & \multicolumn{1}{c}{\textbf{HS-3}} & \multicolumn{1}{c}{\textbf{HS-4}}
         & \multicolumn{1}{c}{\textbf{HS-5}}\\\bottomrule
         SBDCNN & $49.71$& $24.33$& $42.10$ & $41.72$& $56.83$\\ \hline
         VCN~(GPT) & $54.46$& $29.02$& $46.76$ & $48.45$ &$62.81$\\ \hline
         DSTFN & $\mathbf{64.56}$& $\mathbf{31.43}$& $\mathbf{55.99}$& $\mathbf{51.18}$ & $\mathbf{67.38}$\\ \bottomrule 
    \end{tabular}
\end{table}
\begin{table}[]
    \centering
    \caption{Automatic evaluation on the test set under the circumstance that each handwritten Chinese character misses the last stroke.}
    \label{tab:result_5}
    \begin{tabular}{lccccc}
         \multicolumn{1}{c}{\textbf{Model}} & \multicolumn{1}{c}{\textbf{P@1}} & \multicolumn{1}{c}{\textbf{P@2}} & \multicolumn{1}{c}{\textbf{P@3}} & \multicolumn{1}{c}{\textbf{P@4}}
         & \multicolumn{1}{c}{\textbf{P@5}}\\\bottomrule
         SBDCNN & $88.87$& $92.37$& $93.62$ & $94.35$& $94.85$\\ \hline
         VCN~(GPT) & $90.27$& $93.02$& $93.95$ & $94.50$ &$94.89$\\ \hline
         DSTFN & $\mathbf{92.81}$& $\mathbf{95.22}$& $\mathbf{96.01}$& $\mathbf{96.45}$ & $\mathbf{96.72}$\\ \bottomrule 
    \end{tabular}
\end{table}
\begin{table}[]
    \centering
    \caption{Automatic evaluation~(P@1) on the test set with five selected handwritten sets where each handwritten Chinese character misses the last stroke.}
    \label{tab:result_6}
    \begin{tabular}{lccccc}
         \multicolumn{1}{c}{\textbf{Model}} & \multicolumn{1}{c}{\textbf{HS-1}} & \multicolumn{1}{c}{\textbf{HS-2}} & \multicolumn{1}{c}{\textbf{HS-3}} & \multicolumn{1}{c}{\textbf{HS-4}}
         & \multicolumn{1}{c}{\textbf{HS-5}}\\\bottomrule
         SBDCNN & $75.34$& $31.10$& $55.94$ & $58.95$& $78.27$\\ \hline
         VCN~(GPT) & $78.67$& $35.48$& $56.82$ & $63.41$ &$79.73$\\ \hline
         DSTFN & $\mathbf{81.59}$& $\mathbf{37.98}$& $\mathbf{60.95}$& $\mathbf{70.45}$ & $\mathbf{82.00}$\\ \bottomrule 
    \end{tabular}
\end{table}

\subsubsection{\textbf{Robustness and Efficiency}} To evaluate the robustness of models, we test them under the circumstance that each handwritten Chinese character retains 70\% strokes or losses the last stroke. We retrain the models introduced in Section~\ref{experi} under the above circumstances of missing strokes, and the test procedure is identical as in the case of complete strokes except that the stroke retention of handwritten characters is different. The experiment results in~\cref{tab:result_3,tab:result_4,tab:result_5,tab:result_6} show that DSTFN significantly exceeds VCN and SBDCNN on all evaluation metrics~(e.g. P@1, HS-2/3). The results also demonstrate that DSTFN is more robust than the other two models due to less performance degradation. In addition, we observe that the recognition ability of DSTFN decreases less than the other two models as the strokes of handwritten character reduce, especially on P@5. It further indicates that it is a feasible and robust method to integrate the glyph features of handwritten characters and their contextual information multi times based on the strong memory ability of pre-trained autoregressive framework.

\begin{figure}[t]
    \centering
    \includegraphics[width=0.45\textwidth]{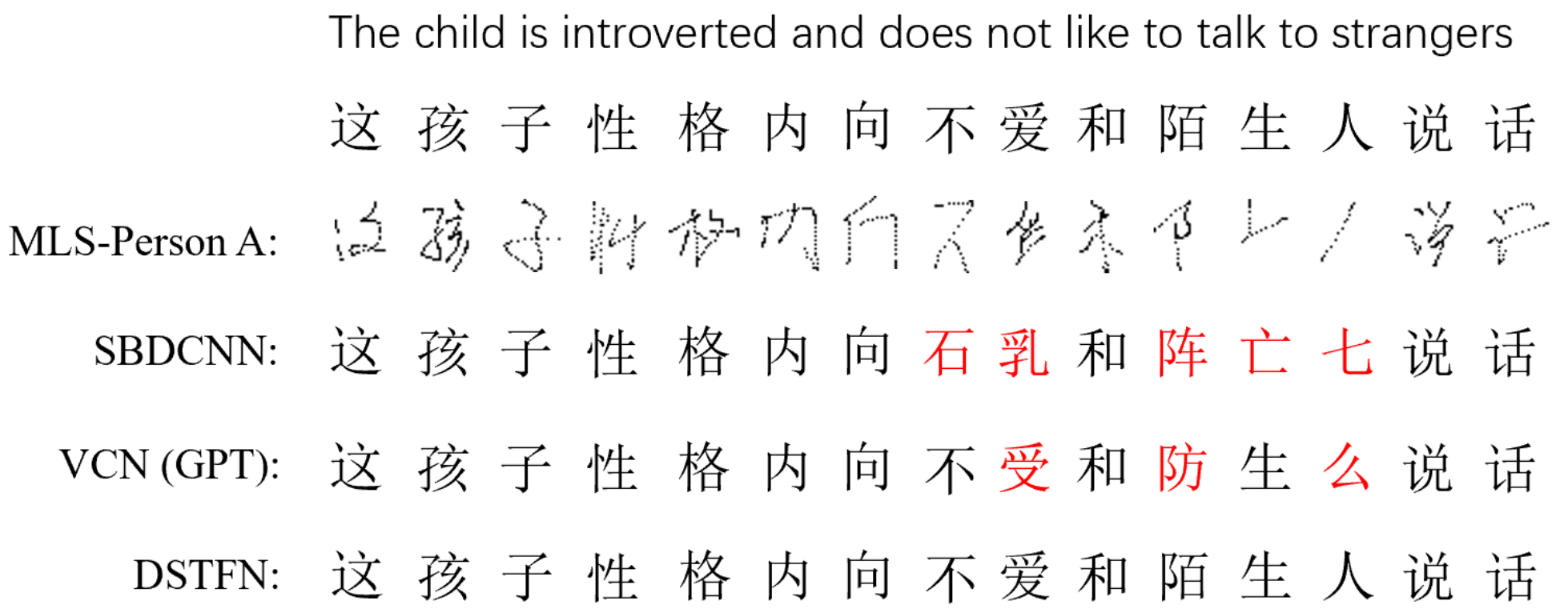}
    \caption{MLS-Person A represents that each handwritten character is with missing the last stroke. Red colored words are the wrong recognition result~(same as the figure below).}
    \label{Fig:instance_2}
\end{figure}
\begin{figure}[H]
    \centering
    \includegraphics[width=0.45\textwidth]{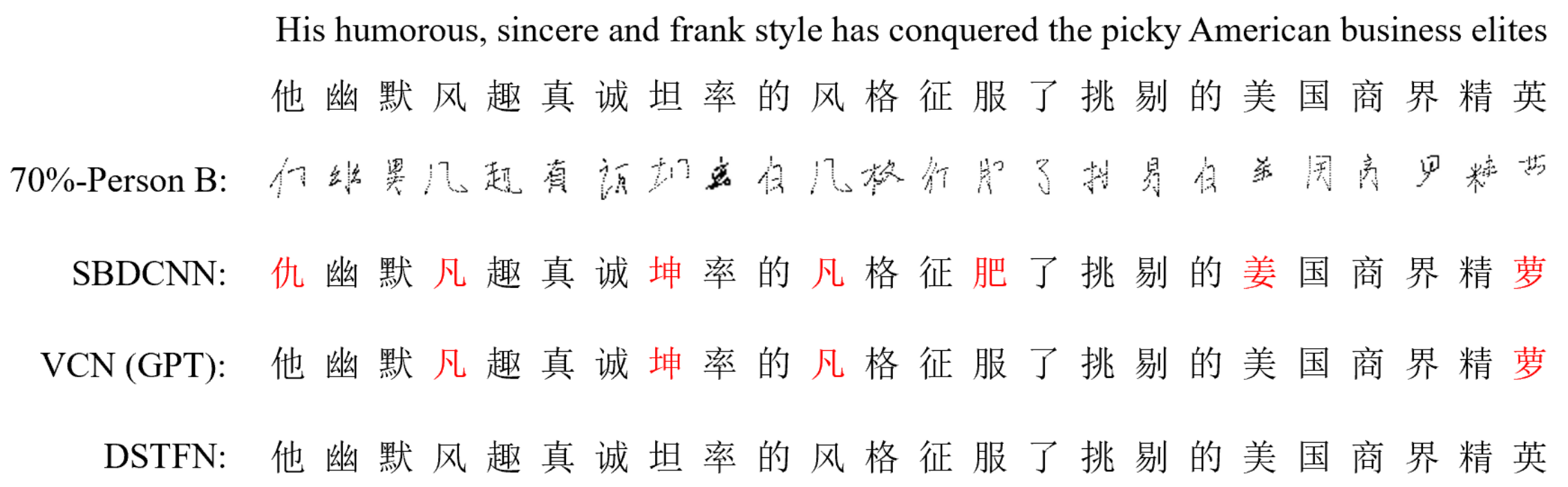}
    \caption{The second recognition instance under the circumstance that each handwritten character is with retaining 70\% of all strokes.}
    \label{Fig:instance_3}
\end{figure}

\begin{CJK}{UTF8}{gbsn}
As shown in Figure~\ref{Fig:instance_2}, Chinese characters with similar structures are extremely identical while missing strokes, such as the third last words of the sentence: '七'~(seven), '么'~(me) and '人'~(people) have the similar stroke '丿'. 
Hence, the spatial features of handwritten characters with missing strokes extracted by the deep convolutional network easily lack the discrimination. It may lead to the week robustness of DSTFN and VCN(GPT). From Figure \ref{Fig:instance_3}, we observe that DSTFN still recognizes '坦'~(smooth) accurately while it retains 70\% of strokes, which has a similar structure with another Chinese character '坤'~(Kun) at this time. It may be attributed to the fact that the phrase '真诚坦率'~(sincere and frank) is often used by persons and DSTFN can learn and utilize words combination information well.\end{CJK}  
From the above experiment results and instances, we observe that DSTFN performs best among all compared models whether it is a simple or complex Chinese character, whether it has complete or incomplete handwritten forms. In conclusion, DSTFN is the most robust model among compared models for recognizing sentence-level online handwriting input.



Improving the recognition accuracy of incomplete handwritten Chinese characters can promote the efficiency of sentence-level handwriting input. Hence, with the exception of analyzing the accuracy and robustness of proposed models in the above experimental cases, we now analyze the efficiency of models by the way that the strokes of handwritten Chinese characters are continuously reduced in sentences. We carefully evaluate the efficiency of models from two aspects. One is the overall recognition accuracy on the test set when strokes of handwritten Chinese characters are continuously decreased as the length of sentence increase; the other is counting how many strokes models use at least to accurately recognize any position in sentences.

\begin{table}[]
    \centering
    \caption{Automatic evaluation on the test set under the circumstance that strokes of handwritten Chinese characters in one sentence are reduced as the length of the sentence.}
    \label{tab:result_7}
    \begin{tabular}{lccccc}
         \multicolumn{1}{c}{\textbf{Model}} & \multicolumn{1}{c}{\textbf{P@1}} & \multicolumn{1}{c}{\textbf{P@2}} & \multicolumn{1}{c}{\textbf{P@3}} & \multicolumn{1}{c}{\textbf{P@4}}
         & \multicolumn{1}{c}{\textbf{P@5}}\\\bottomrule
         SBDCNN & $71.41$& $80.76$& $84.58$ & $86.93$& $88.55$\\ \hline
         VCN~(GPT) & $84.07$& $90.79$& $93.13$ & $94.33$ &$95.22$\\ \hline
         DSTFN & $\mathbf{90.68}$& $\mathbf{94.61}$& $\mathbf{95.93}$& $\mathbf{96.66}$ & $\mathbf{97.14}$\\ \bottomrule 
    \end{tabular}
\end{table}
\begin{table}[]
    \centering
    \caption{Automatic evaluation~(P@1) on the test set with five selected handwritten sets.}
    \label{tab:result_8}
    \begin{tabular}{lccccc}
         \multicolumn{1}{c}{\textbf{Model}} & \multicolumn{1}{c}{\textbf{HS-1}} & \multicolumn{1}{c}{\textbf{HS-2}} & \multicolumn{1}{c}{\textbf{HS-3}} & \multicolumn{1}{c}{\textbf{HS-4}}
         & \multicolumn{1}{c}{\textbf{HS-5}}\\\bottomrule
         SBDCNN & $67.38$& $40.63$& $53.25$ & $55.41$& $65.16$\\ \hline
         VCN~(GPT) & $78.40$& $51.64$ & $66.21$ & $68.54$ &$78.53$\\ \hline
         DSTFN & $\mathbf{85.41}$ & $\mathbf{57.29}$& $\mathbf{73.78}$& $\mathbf{79.66}$ & $\mathbf{86.77}$\\ \bottomrule 
    \end{tabular}
\end{table}

Firstly, we train models on the training data under the circumstance that handwritten Chinese characters in different positions retain different numbers of strokes with different probabilities. That is, the characters in the first 25\% of the sentence retain the complete stroke or only miss the last stroke with a higher probability; the Chinese characters in the 25\% $\sim{}$50\% positions retain 70\% of strokes with a higher probability; the characters in the 50\% $\sim{}$75\% positions retain 50\% of strokes with a higher probability; and Chinese characters in the last part retain 30\% of strokes with a higher probability. More detailed experiment settings are shown in Appendix A. During inferring, for the first 75\% positions of the sentence, we reduce one stroke of the handwritten Chinese character according to 25\% intervals, and retain 70\% of strokes for the characters in the last 25\% part. We test models on the test set five times and the average results are shown in Table~\ref{tab:result_7}. The recognition accuracy of the selected five sets is shown in Table~\ref{tab:result_8}.

Compared to the circumstance of complete strokes, the experiment results of Table~\ref{tab:result_7} and Table~\ref{tab:result_8} show that the performance of all models drop drastically on P@1, yet DSTFN can achieve approximately 97\% accuracy on P@5. It shows that DSTFN can maintain high recognition accuracy for each character in the case that the strokes of Chinese characters are continuously reduced. The above phenomenon indicates DSTFN can promote the efficiency of sentence-level handwriting input by the above way.
According to the intrinsic step-by-step prediction way of sequential classification, the high recognition accuracy of p@1 can further promote the overall accuracy and efficiency at inferring sentence-level handwriting input. It can reduce the cascading errors. Hence, it is challenging to promote the efficiency of sentence-level handwriting input, judging from the p@1 evaluation results. From Table~\ref{tab:result_8} and Figure~\ref{fig:stroke_two}, we can observe that DSTFN can not handle the extremely irregular and sloppy handwritten characters well under the circumstance that strokes of characters are continuously reduced in a sentence. The above analyses indicate that DSTFN is enough robust and efficient to be used in normal sentence-level handwriting scenarios, yet the accuracy and robustness still have a lot of room for improvement while confronting not well written characters.      

\begin{figure}[t]
    \centering
    \includegraphics[width=0.45\textwidth]{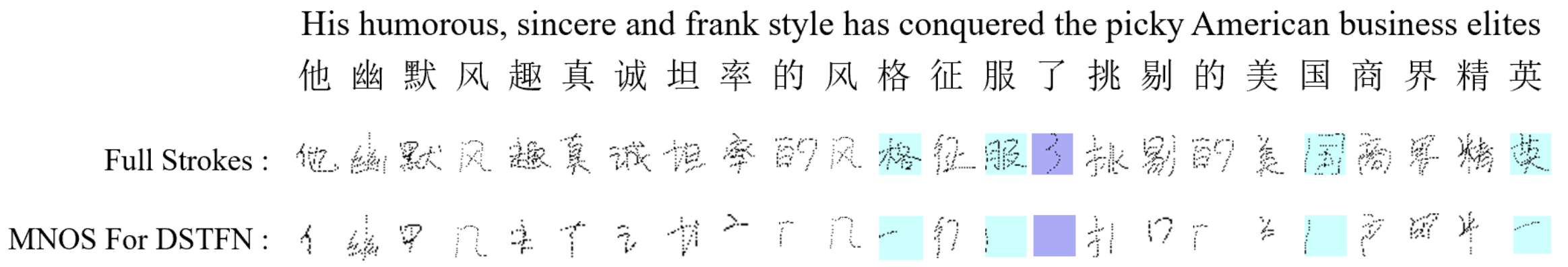}
    \caption{An instance of correct recognition under the minimum number of strokes. MNOS FOR DSTFN is the minimum number of strokes for DSTFN. The purple box represents that it can still be recognized accurately even if there are no written strokes sometimes. The light blue boxes represent that DSTFN can accurately recognize Chinese characters written with only one stroke.}
    \label{fig:last_s}
\end{figure}

\begin{figure}[t]
    \centering
    \includegraphics[width=0.45\textwidth]{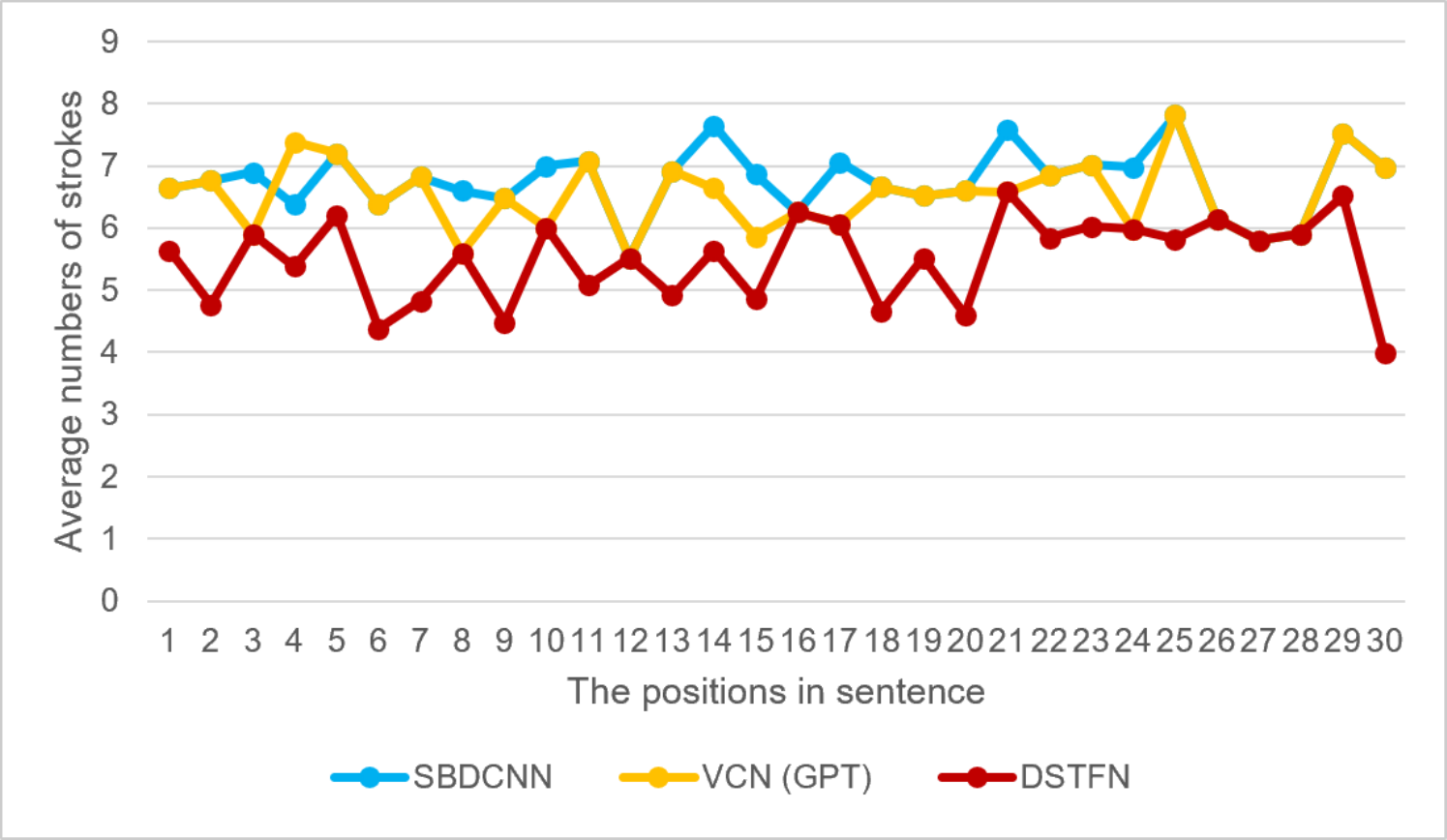}
    \caption{The statistics of minimum number of strokes at each position recognized accurately by models. The evaluation metric of accuracy is P@3.}
    \label{fig:last}
\end{figure}

Secondly, we select 150 sentences with a length of 30 words, and count the minimum number of strokes at each position used to be recognized accurately by models. From the statistic results shown in Figure~\ref{fig:last}, we observe that each line fluctuates up and down, which is caused by characters at different positions having different numbers of strokes. Moreover, the ups and downs of the DSTFN curve are related to the learned short-distance dependency of phrases in Chinese sentences. As shown in Figure~\ref{fig:last_s}, many characters only need at most one stroke to be recognized accurately by DSTFN. Hence, Compared with the SBDCNN and VCN~(GPT), DSTFN can recognize handwritten Chinese characters with fewer strokes precisely in most positions, which further promote the recognition efficiency of it.

\section{Conclusion}

In this paper, we explore how to design precise, robust and effective methods for sentence-level online handwritten Chinese character recognition~(sentence-level OLHCCR). First, we construct Chinese sentence-level online handwriting dataset named as CSOHD to step forward the research of sentence-level OLHCCR. Second, we propose the vanilla compositional network~(VCN) coupled deep convolutional network with language model, and further design deep spatial-temporal fusion network~(DSTFN) in view of not well written characters. Finally, we conduct extensive experiments to evaluate the accuracy, robustness and efficiency of our models. DSTFN performs best and achieves excellent performance under all scenarios. As for implementation, DSTFN with large parameters can be applied in actual applications efficiently with the rapid development of large-scale cloud servers.  

\bibliographystyle{ACM-Reference-Format}
\bibliography{sample-authordraft}


\end{document}







\maketitle
\appendix
\section{Experiment Setup}

\subsection{Training Method}

In this section, we introduce the concrete training method we adopt in Section 5.2.2. In order to be adapt to the continuous reduction scenario of strokes, we adjust the number of handwritten character strokes retained in different positions of the sentence in the form of probability during training. As shown in Table~\ref{tab:propo_train}, We first let the model learn while retaining more strokes at different positions and further let the model further learn with retaining fewer strokes as shown in Figure~\ref{tab:propo_train_2}. It makes models learn from simple to difficult in the way of course learning.
\begin{table}[H]
\renewcommand\arraystretch{1.25}
    \centering
    \small
    \caption{Different positions retain different proportions of stokes during the first training period. MLS represents missing the last stroke. 70\%, 50\% and 30\% represent retraining the corresponding proportion of strokes for each handwritten Chinese character. The content of the table is the proportion of different numbers of strokes selected for each position, and the sum of the probabilities of each row is 1.}
   \label{tab:propo_train}
   \begin{tabular}{c|ccccc}
        \hline
        \diagbox{Position}{Retaining} & \multicolumn{1}{c}{Full} & \multicolumn{1}{c}{MLS} & \multicolumn{1}{c}{70\%} & \multicolumn{1}{c}{50\%} & \multicolumn{1}{c}{30\%}\\
        \hline
        \textbf{\textbf{0 $\sim{}$25\%}} & 60\% & 30\% & 10\% & 0 & 0\\
        \textbf{25\% $\sim{}$50\%} & 10\% & 40\% & 40\% & 10\% & 0\\
        \textbf{50\% $\sim{}$75\%} & 10\% & 10\% & 30\% & 40\% & 10\%\\
        \textbf{75\% $\sim{}$100\%} & 10\% & 10\% & 20\% & 40\% & 20\%\\
        \hline
    \end{tabular}
\end{table}
\begin{table}[H]
\renewcommand\arraystretch{1.25}
    \centering
    \small
    \caption{Different positions retain different proportions of stokes during the second training period.}
   \label{tab:propo_train_2}
   \begin{tabular}{c|ccccc}
        \hline
        \diagbox{Position}{Retaining} & \multicolumn{1}{c}{Full} & \multicolumn{1}{c}{MLS} & \multicolumn{1}{c}{70\%} & \multicolumn{1}{c}{50\%} & \multicolumn{1}{c}{30\%}\\
        \hline
        \textbf{\textbf{0 $\sim{}$25\%}} & 5\% & 45\% & 50\% & 0 & 0\\
        \textbf{25\% $\sim{}$50\%} & 0 & 30\% & 40\% & 30\% & 0\\
        \textbf{50\% $\sim{}$75\%} & 0 & 10\% & 20\% & 40\% & 30\%\\
        \textbf{75\% $\sim{}$100\%} & 0 & 0 & 20\% & 30\% & 50\%\\
        \hline
    \end{tabular}
\end{table}